# Formula-Based Probabilistic Inference

**Vibhav Gogate** and **Pedro Domingos**
Computer Science & Engineering,
University of Washington,
Seattle, WA 98195, USA.
{vgogate,pedrod}@cs.washington.edu

## Abstract

Computing the probability of a formula given the probabilities or weights associated with other formulas is a natural extension of logical inference to the probabilistic setting. Surprisingly, this problem has received little attention in the literature to date, particularly considering that it includes many standard inference problems as special cases. In this paper, we propose two algorithms for this problem: *formula decomposition and conditioning*, which is an exact method, and *formula importance sampling*, which is an approximate method. The latter is, to our knowledge, the first application of model counting to approximate probabilistic inference. Unlike conventional variable-based algorithms, our algorithms work in the *dual* realm of logical formulas. Theoretically, we show that our algorithms can greatly improve efficiency by exploiting the structural information in the formulas. Empirically, we show that they are indeed quite powerful, often achieving substantial performance gains over state-of-the-art schemes.

## 1 Introduction

The standard task in the field of automated reasoning is to determine whether a set of logical formulas (the knowledge base $KB$) entails a query formula $Q$. (The formulas could be propositional or first-order; in this paper we focus on the propositional case.) Logic's lack of a representation for uncertainty severely hinders its ability to model real applications, and thus many methods for adding probability to it have been proposed. One of the earliest is Nilsson's probabilistic logic (Nilsson, 1986), which attaches probabilities to the formulas in the KB and uses these to compute the probability of the query formula. One problem with this approach is that the formula probabilities may be inconsistent, yielding no solution, but consistency can be verified and enforced (Nilsson, 1986). Another problem is that in general a set of formula probabilities does not completely specify a distribution, but this is naturally solved by assuming the maximum entropy distribution consistent with the specified probabilities (Nilsson, 1986; Pietra et al., 1997).

A more serious problem is the lack of efficient inference procedures for probabilistic logic. This contrasts with the large literature on inference for graphical models, which always specify unique and consistent distributions (Pearl, 1988). However, the representational flexibility and compactness of logic is highly desirable, particularly for modeling complex domains. This issue has gained prominence in the field of statistical relational learning (SRL) (Getoor and Taskar, 2007), which seeks to learn models with both logical and probabilistic aspects. For example, Markov logic represents knowledge as a set of weighted formulas, which define a log-linear model (Domingos and Lowd, 2009). Formulas with probabilities with the maximum entropy assumption and weighted formulas are equivalent; the problem of converting the former to the latter is equivalent to the problem of learning the maximum likelihood weights (Pietra et al., 1997). In this paper we assume weighted formulas, but our algorithms are applicable to formulas with probabilities by first performing this conversion.

Another reason to seek efficient inference procedures for probabilistic logic is that inference in graphical models can be reduced to it (Park, 2002). Standard inference schemes for graphical models such as junction trees (Lauritzen and Spiegelhalter, 1988) and bucket elimination (Dechter, 1999) have complexity exponential in the treewidth of the model, making them impractical for complex domains. However, treewidth can be overcome by exploiting structural properties like determinism (Chavira and Darwiche, 2008) and context-specific independence (Boutilier, 1996). Several highly efficient algorithms accomplish this by encoding a graphical models as sets of weighted formulas and applying logical inference techniques to them (Sang et al., 2005; Chavira and Darwiche, 2008).

All of these algorithms are *variable-based*, in that they explore the search space defined by truth assignments to the variables. In this paper, we propose a new class of algorithms that explore the search space defined by truth assignments to arbitrary formulas, including but not necessarily those contained in the original specification. Our *formula-based* schemes generalize variable-based schemes because a variable is a special case of a formula, namely a unit clause. For deriving exact answers, we propose to exhaustively search the space of truth assignments to formulas, yielding the formula decomposition and conditioning (FDC) scheme. FDC performs AND/OR search (Dechter and Mateescu, 2007) or recursive conditioning (Darwiche, 2001), with and without caching, over the space of formulas, utilizing several Boolean constraint propagation and pruning techniques.

Even with these techniques, large complex domains will still generally require approximate inference. For this, we propose to compute an importance distribution over the formulas, yielding formula importance sampling (FIS). Each sample in FIS is a truth assignment to a set of formulas. To compute the importance weight of each such sampled assignment, we need to know its model count (or number of solutions). These model counts can either be computed exactly, if it is feasible, or approximately using the recently introduced approximate model counters such as Sample-Count (Gomes et al., 2007) and SampleSearch (Gogate and Dechter, 2007b). To the best of our knowledge, this is the first work that harnesses the power of model counting for approximate probabilistic inference. We prove that if the model counts can be computed accurately, formula importance sampling will have smaller variance than variable-based importance sampling and thus should be preferred.

We present experimental results on three classes of benchmark problems: random Markov networks, QMR-DT networks from the medical diagnosis domain and Markov logic networks. Our experiments show that as the number of variables in the formulas increases, formula-based schemes not only dominate their variable based counterparts but also state-of-the-art exact algorithms such as ACE (Chavira and Darwiche, 2008) and approximate schemes such as MC-SAT (Poon and Domingos, 2006) and Gibbs sampling (Geman and Geman, 1984).

The rest of the paper is organized as follows. Section 2 describes background. Section 3 presents formula decomposition and conditioning. Section 4 presents formula importance sampling. Experimental results are presented in Section 5 and we conclude in Section 6.

## 2 Background

### 2.1 Notation

Let $\mathbf{X} = \{X_1, \ldots, X_n\}$ be a set of propositional variables that can be assigned values from the set $\{0, 1\}$ or $\{False, True\}$. Let $F$ be a propositional formula over $\mathbf{X}$.

A model or a solution of $F$ is a $0/1$ truth assignment to all variables in $\mathbf{X}$ such that $F$ evaluates to True. We will assume throughout that $F$ is in CNF, namely it is a conjunction of clauses, a clause being a disjunction of literals. A literal is a variable $X_i$ or its negation $\neg X_i$. A unit clause is a clause with one literal. Propositional Satisfiability or SAT is the decision problem of determining whether $F$ has any models. This is the canonical NP-complete problem. Model Counting is the problem of determining the number of models of $F$, it is a #P-complete problem.

We will denote formulas by letters $F$, $G$, and $H$, the set of solutions of $F$ by $Sol(F)$ and its number of solutions by $\#(F)$. Variables are denoted by letters $X$ and $Y$. We denote sets by bold capital letters e.g., $\mathbf{X}$, $\mathbf{Y}$ etc. Given a set $\mathbf{X} = \{X_1, \ldots, X_n\}$ of variables, $\mathbf{x}$ denotes a truth assignment $(x_1, \ldots, x_n)$, where $X_i$ is assigned the value $x_i$. Clauses are denoted by the letters $C$, $R$, $S$ and $T$. Discrete functions are denoted by small Greek letters, e.g. $\phi$, $\psi$, etc. The variables involved in a function $\phi$, namely the scope of $\phi$ is denoted by $V(\phi)$. Similarly, the variables of a clause $C$ are denoted by $V(C)$. Given an assignment $\mathbf{x}$ to a superset $\mathbf{X}$ of $\mathbf{Y}$, $\mathbf{x_Y}$ denotes the restriction of $\mathbf{x}$ to $\mathbf{Y}$.

The expected value of a random variable $X$ with respect to a distribution $Q$ is $\mathbb{E}_Q[X] = \sum_{x \in X} x Q(x)$. The variance of $x$ is $Var_Q[X] = \sum_{x \in X} (x - \mathbb{E}_Q[X])^2 Q(x)$.

In this paper, we advocate using a collection of weighted propositional formulas instead of the conventional tabular representations to encode the potentials in Markov random fields (MRFs) or conditional probability tables in Bayesian networks. Specifically, we will use the following representation, which we call as propositional MRF or `PropMRF` in short. A `PropMRF` is a Markov logic network (Richardson and Domingos, 2006) in which all formulas are propositional. It is known that any discrete Markov random field or a Bayesian network can be encoded as a `PropMRF` (Park, 2002; Sang et al., 2005; Chavira and Darwiche, 2008).

DEFINITION 1 (**Propositional MRFs**). A propositional MRF (`PropMRF`), denoted by $\mathcal{M}$ is a triple $(\mathbf{X}, \mathbf{C}, \mathbf{R})$ where $\mathbf{X}$ is a set of $n$ Boolean variables, $\mathbf{C} = \{(C_1, w_1), \ldots, (C_m, w_m)\}$ is a set of $m$ soft (weighted) clauses and $\mathbf{R} = \{R_1, \ldots, R_p\}$ is a set of $p$ hard clauses. Each soft clause is a pair $(C_i, w_i)$ where $C_i$ is a clause and $w_i$ is a real number. We will denote by $F_\mathcal{M} = R_1 \wedge \ldots \wedge R_p$, the CNF formula defined by the hard clauses of $\mathcal{M}$. The **primal graph** of a `PropMRF` has variables as its vertices and an edge between any two nodes that are involved in the same hard or soft clause.

We can associate a discrete function $\phi_i$ with each soft clause $(C_i, w_i)$, defined as follows:

$$\phi_i(\mathbf{x}_{V(C_i)}) = \begin{cases} \exp(w_i) & \text{If } \mathbf{x} \text{ evaluates } C_i \text{ to True} \\ 1 & \text{Otherwise} \end{cases}$$

The probability distribution associated with $\mathcal{M}$ is given by:

$$P_\mathcal{M}(\mathbf{x}) = \begin{cases} \frac{1}{Z_\mathcal{M}} \prod_{i=1}^m \phi_i(\mathbf{x}_{V(\phi_i)}) & \text{If } \mathbf{x} \in Sol(F_\mathcal{M}) \\ 0 & \text{Otherwise} \end{cases} \quad (1)$$

where $Z_\mathcal{M}$ is the normalization constant; often referred to as the *partition function*. $Z_\mathcal{M}$ is given by:

$$Z_\mathcal{M} = \sum_{\mathbf{x} \in Sol(F_\mathcal{M})} \prod_{i=1}^m \phi_i(\mathbf{x}_{V(\phi_i)}) \quad (2)$$

Note that if $\mathcal{M}$ has no soft clauses, then $Z_\mathcal{M}$ equals the number of models of the formula $F_\mathcal{M}$. Thus, model counting is a special case of computing $Z_\mathcal{M}$.

We will focus on the query of finding the probability of a CNF formula $G$, denoted by $P(G)$. By definition:

$$\begin{align} P(G) &= \sum_{\mathbf{x} \in Sol(F_\mathcal{M} \wedge G)} P_\mathcal{M}(\mathbf{x}) \\ &= \frac{1}{Z} \sum_{\mathbf{x} \in Sol(F_\mathcal{M} \wedge G)} \prod_{i=1}^m \phi_i(\mathbf{x}_{V(\phi_i)}) \end{align} \quad (3)$$

If we add all the clauses of $G$ to the hard clauses of $\mathcal{M}$ yielding another PropMRF $\mathcal{M}'$, then the partition function $Z_{\mathcal{M}'}$ of $\mathcal{M}'$ is given by:

$$Z_{\mathcal{M}'} = \sum_{\mathbf{x} \in Sol(F_\mathcal{M} \wedge G)} \prod_{i=1}^m \phi_i(\mathbf{x}_{V(\phi_i)}) \quad (4)$$

From Equations 3 and 4, we get $P(G) = \frac{Z_{\mathcal{M}'}}{Z_\mathcal{M}}$. Because computing $P(G)$ is equivalent to computing a ratio of two partition functions, in the sequel, we will present *formula-based algorithms* for computing $Z_\mathcal{M}$ only.

## 3 Exact Formula-based Inference

We first explain how to perform inference by variable-based conditioning and then show how it can be generalized via formula-based conditioning. Consider the expression for $Z_\mathcal{M}$ (See Equation 2). Given assignments $X_j$ and $\neg X_j$, we can express $Z_\mathcal{M}$ as:

$$\begin{align} Z_\mathcal{M} &= \sum_{\mathbf{x} \in Sol(F_\mathcal{M} \wedge X_j)} \prod_{i=1}^m \phi_i(\mathbf{x}_{V(\phi_i)}) \\ &\quad + \sum_{\mathbf{x} \in Sol(F_\mathcal{M} \wedge \neg X_j)} \prod_{i=1}^m \phi_i(\mathbf{x}_{V(\phi_i)}) \quad (5) \\ &= Z_{\mathcal{M}_{X_j}} + Z_{\mathcal{M}_{\neg X_j}} \quad (6) \end{align}$$

where $\mathcal{M}_X$ and $\mathcal{M}_{\neg X}$ are PropMRFs obtained by adding $X$ and $\neg X$ to the set of hard clauses of $\mathcal{M}$ respectively.

Then, one can perform conditioning to compute $Z_{\mathcal{M}_{X_j}}$ and $Z_{\mathcal{M}_{\neg X_j}}$, recursively for each PropMRF until all variables in $\mathbf{X}$ have been instantiated. Conditioning by itself is not that useful. For example, if the PropMRF has no hard clauses, then conditioning would perform $2^n$ summations. However, one can augment it with various simplification schemes such as Boolean constraint propagation, and utilize problem decomposition, yielding powerful schemes in practice. These and other ideas form the backbone of many state-of-the-art schemes such as ACE (Chavira and Darwiche, 2008) and Cachet (Sang et al., 2005). To simplify a PropMRF, we can apply any parsimonious operators - operators which do not change its partition function. In particular, we can remove all clauses which evaluate to True from the set of hard clauses. These clauses are redundant. Examples of operations that aid in identifying such hard clauses are unit propagation, resolution and subsumption elimination. For example, given a hard clause $A$, the hard clause $A \vee B$ is redundant and can be removed because it is subsumed within $A$. Similarly, $A$ could be removed after unit propagation, because it always evaluates to True. We can simplify the soft clauses based on the hard clauses by removing all soft clauses which evaluate to either True or False, multiplying the partition function with an appropriate constant to account for their removal. For example, given a hard clause $A$, the soft clause $A \vee B$ having weight $w$ is always satisfied and can be removed, by multiplying the partition function by $\exp(w)$[1].

Another advancement that we can use is problem decomposition (Darwiche, 2001; Dechter and Mateescu, 2007). The idea here is that if the soft and hard clauses of a PropMRF can be partitioned into $k > 1$ sets such that any two clauses in any of the $k$ sets have no variables in com-mon, then the partition function equals the product of the partition functions of the $k$ `PropMRF`s induced by each set.

| ClauseID | Clause | weight |
|---|---|---|
| S1 | $A \vee B \vee C \vee D \vee E$ | $w_1$ |
| S2 | $A \vee B \vee C \vee F \vee G$ | $w_2$ |
| S3 | $D \vee E \vee H$ | $w_3$ |
| S4 | $F \vee G \vee J$ | $w_4$ |

Figure 1: An example PropMRF.

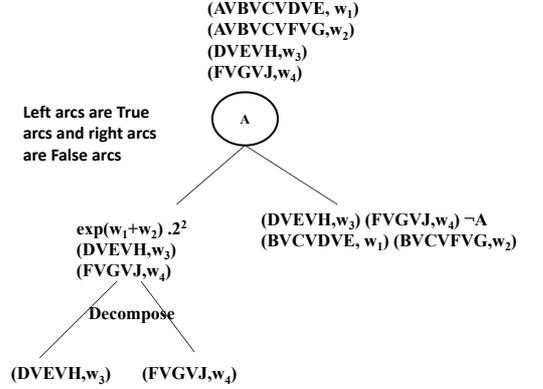

Figure 2: Figure demonstrating the simplification steps after conditioning on variable A for the PropMRFgiven in Figure 1.

---

[1] Note that if we remove a variable that is not a unit clause from all the hard and soft clauses, then we have to multiply the partition function by 2.

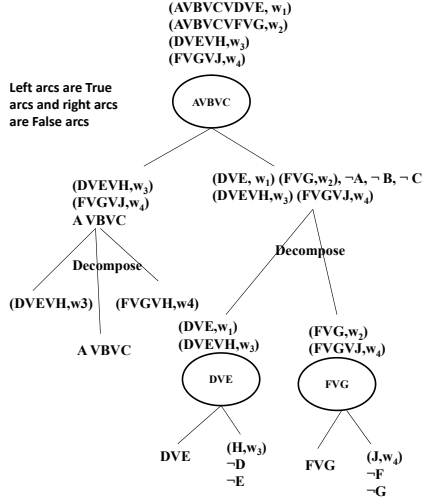

Figure 3: Search space of Formula Decomposition and Conditioning for an example `PropMRF`

The following example demonstrates simplification and decomposition on an example `PropMRF`.

EXAMPLE 1. Consider the `PropMRF` shown in Figure 1. After conditioning on $A$ and simplifying using Boolean constraint propagation, we get two `PropMRF`s shown under the True (left) and false (right) branches of $A$ in Figure 2. The `PropMRF` at the True branch contains only two soft clauses which have no variables in common. Thus, they could be decomposed into two `PropMRF`s as shown. The contribution to the partition function due to the True branch of $A$ is then simply a product of the partition functions of the two `PropMRF`s and $\exp(w_1 + w_2) \times 2^2$.

Our main observation is that we can condition on arbitrary formulas instead of variables. Formally, given an arbitrary formula $H_j$, we can express $Z_\mathcal{M}$ as:

$$\begin{aligned}
Z_\mathcal{M} &= \sum_{\mathbf{x} \in Sol(F_\mathcal{M} \wedge H_j)} \prod_{i=1}^{m} \phi_i(\mathbf{x}_{V(\phi_i)}) \\
&+ \sum_{\mathbf{x} \in Sol(F_\mathcal{M} \wedge \neg H_j)} \prod_{i=1}^{m} \phi_i(\mathbf{x}_{V(\phi_i)}) \quad (7) \\
&= Z_{\mathcal{M}_{H_j}} + Z_{\mathcal{M}_{\neg H_j}} \quad (8)
\end{aligned}$$

When combined with Boolean constraint propagation and problem decomposition, this seemingly simple idea is quite powerful because it can yield a smaller search space, as we demonstrate in the following example. In some cases, these reductions could be significant.

EXAMPLE 2. Consider again the `PropMRF` shown in Figure 1. The first two clauses share a sub-clause $A \vee B \vee C$. If we condition first on $A \vee B \vee C$, we get the search space shown in Figure 3, which has only 7 leaf nodes. One can

**Algorithm 1**: Formula Decomposition and Conditioning (FDC)

**Input**: A `PropMRF` $\mathcal{M}$
**Output**: $Z_\mathcal{M}$
**begin**
　　$w = 0$;
　　**1. Simplify**
　　**begin**
　　　　Simplify the hard and soft clauses;
　　　　Add the weights of all soft clauses which evaluate to True to $w$;
　　　　Remove all soft clauses which evaluate to either True or False from $\mathcal{M}$. Update $w$ to account for variables completely removed from all formulas;
　　　　**if** $F_\mathcal{M}$ *has an empty clause* **then**
　　　　　　**return** 0
　　　　**if** $F_\mathcal{M}$ *has only unit clauses* **then**
　　　　　　**return** $\exp(w)$
　　**end**
　　**2. Decompose**
　　**begin**
　　　　**if** *the primal graph of $\mathcal{M}$ is decomposable into $k$ components* **then**
　　　　　　Let $\mathcal{M}_1, \mathcal{M}_2, ..., \mathcal{M}_k$ be the `PropMRF`'s corresponding to the $k$ components;
　　　　　　**return** $\exp(w) \times FDC(\mathcal{M}_1) \times \ldots \times FDC(\mathcal{M}_k)$
　　**end**
　　**3. Condition**
　　**begin**
　　　　Heuristically choose a formula $R$ to condition on;
　　　　Add hard clauses logically equivalent to $R$ and $\neg R$ to $\mathcal{M}$ yielding $\mathcal{M}_R$ and $\mathcal{M}_{\neg R}$ respectively;
　　　　**return** $\exp(w) \times (FDC(\mathcal{M}_R) + FDC(\mathcal{M}_{\neg R}))$
　　**end**
**end**

verify that if we condition only on the variables instead of arbitrary formulas, the best ordering scheme will explore 12 leaf nodes. (This search space is not shown because of lack of space. It can be worked out using Figure 2.)

Algorithm Formula Decomposition and Conditioning (FDC) is presented as Algorithm 1. It takes as input a `PropMRF` $\mathcal{M}$. The first step is the simplification step in which we reduce the size of the hard and the soft clauses using techniques such as unit propagation, resolution and subsumption elimination. In the decomposition step (Step 2), we decompose $\mathcal{M}$ into independent `PropMRF`s if its primal graph is decomposable. Each of them are then solved independently. Note that this is a very important step and is the primary reason for efficiency of techniques such as recursive conditioning (Darwiche, 2001) and AND/OR search (Dechter and Mateescu, 2007). In fact, the whole idea in performing simplification and heuristic conditioning is to split the `PropMRF` into several `PropMRF`'s that can be solved independently. Finally, in the conditioning step, we heuristically select a formula $R$ to condition on and then recurse on the true and the false assignments to $R$.

We summarize the dominance of FDC over VDC (where VDC is same as FDC except that we condition only on unit clauses in Step 3) in the following proposition.

PROPOSITION 1. *Given a* `PropMRF` *$\mathcal{M}$, let $S_{\mathcal{M},F}$ and $S_{\mathcal{M},V}$ be the number of nodes in the smallest search space explored by FDC and VDC respectively. Then $S_{\mathcal{M},F} \leq S_{\mathcal{M},V}$. Sometimes, this inequality can be strict.*

**Improvements**

We consider two important improvements. First, note that if we are not careful, the algorithm as presented may yield a super-exponential search space. For example, if we condition on a set of arbitrary formulas, none of which simplify the `PropMRF`, we may end up conditioning on a super-exponential number of formulas. Trivially, to guarantee at least an exponential search space in the size of the clausal specification, the formula selected for conditioning must reduce/simplify at least one soft clause or at least one hard clause. Second, we can augment FDC with *component caching and clause learning* as in Cachet (Sang et al., 2005) and use *w-cutset conditioning* (Dechter, 1999) in a straight forward manner. We omit the details.

### 3.1 Related work

FDC generalizes variable-based conditioning schemes such as recursive conditioning (Darwiche, 2001), AND/OR search (Dechter and Mateescu, 2007) and value elimination (Bacchus et al., 2003) because all we have to do is restrict our conditioning to unit clauses. FDC also generalizes weighted model counting (WMC) approaches such as ACE (Chavira and Darwiche, 2008) and Cachet (Sang et al., 2005). These weighted model counting approaches introduce additional Boolean variables to model each soft clause. Conditioning on these Boolean variables is equivalent to conditioning on the soft clauses present in the `PropMRF`. Thus, FDC can simulate WMC by restricting its conditioning to not only the unit clauses but also the soft clauses already present in the `PropMRF`. Finally, FDC is related to streamlined constraint reasoning (SCR) approach of (Gomes and Sellmann, 2004). The idea in SCR is to add a set of *streamlining formulas* to the input formula in order to cut down the size of its solution space in a controlled manner. The goal of streamlining is solving a Boolean Satisfiability (or a Constraint Satisfaction) problem while FDC uses (streamlined) formulas for weighted model counting.

## 4 Formula Importance Sampling

In this section, we generalize conventional variable-based importance sampling to formula importance sampling and show that our generalization yields new sampling schemes having smaller variance. We first present background on variable-based importance sampling.

### 4.1 Variable-Based Importance Sampling

Importance sampling (Rubinstein, 1981) is a general scheme which can be used to approximate any quantity such as $Z_\mathcal{M}$ which can be expressed as a sum of a function over a domain. The main idea is to use an importance distribution $Q$, which satisfies $P_\mathcal{M}(\mathbf{x}) > 0 \Rightarrow Q(\mathbf{x}) > 0$ and express $Z_\mathcal{M}$ as follows:

$$\begin{aligned} Z_\mathcal{M} &= \sum_{\mathbf{x} \in Sol(F)} \prod_{i=1}^{m} \phi_i(\mathbf{x}_{V(\phi_i)}) \times \frac{Q(\mathbf{x})}{Q(\mathbf{x})} \\ &= \mathbb{E}_Q \left[ \frac{I(\mathbf{x}) \prod_{i=1}^{m} \phi_i(\mathbf{x}_{V(\phi_i)})}{Q(\mathbf{x})} \right] \end{aligned} \quad (9)$$

where $I(\mathbf{x})$ is an indicator function which is 1 if $\mathbf{x}$ is a solution of $F_\mathcal{M}$ and 0 otherwise.

Given $N$ independent and identical (i.i.d.) samples $(\mathbf{x}^{(1)}, \ldots, \mathbf{x}^{(N)})$ drawn from $Q$, we can estimate $Z_\mathcal{M}$ using $\widehat{Z}_N$, defined below:

$$\widehat{Z}_N = \frac{1}{N} \sum_{i=1}^{N} \frac{I(\mathbf{x}^{(i)}) \prod_{j=1}^{m} \phi_j(\mathbf{x}^{(i)}_{V(\phi_j)})}{Q(\mathbf{x}^{(i)})} \quad (10)$$

It is known (Rubinstein, 1981) that $\mathbb{E}_Q[\widehat{Z}_N] = Z_\mathcal{M}$, namely it is unbiased. The mean squared error (MSE) of $\widehat{Z}_N$ is given by:

$$MSE(\widehat{Z}_N) = \frac{Var_Q \left[ \frac{I(\mathbf{x}) \prod_{i=1}^{m} \phi_i(\mathbf{x}_{V(\phi_i)})}{Q(\mathbf{x})} \right]}{N} \quad (11)$$

Thus, we can reduce the mean squared error by either reducing the variance (given in the numerator) or by increasing the number of samples $N$ (or both).

### 4.2 Formula-based Importance Sampling

Importance sampling can be extended to the space of clauses (or formulas) in a straight forward manner. Let $\mathbf{H} = \{H_1, \ldots, H_r\}$ be a set of arbitrary formulas over the variables $\mathbf{X}$ of $\mathcal{M}$, and let $\mathbf{h} = (h_1, \ldots, h_r)$ be a truth assignment to all the clauses in $\mathbf{H}$. Let $\mathbf{H}$ be such that every consistent truth assignment $\mathbf{h}$ evaluates all soft clauses to either True or False. Note that this condition is critical. Trivially, if $\mathbf{H}$ equals the set of soft clauses, then the condition is satisfied. Let $F_\mathbf{h}$ be the formula corresponding to conjunction $(H_1 = h_1 \wedge \ldots \wedge H_r = h_r)$ and let $\mathbf{x_h} \in Sol(F_\mathbf{h})$. Given a function $\phi$, let $\mathbf{x}_{\mathbf{h},V(\phi)}$ be the restriction of $\mathbf{x_h}$ to the scope of $\phi$. Then, given an importance distribution $U(\mathbf{H})$, we can rewrite $Z_\mathcal{M}$ as:

$$\begin{aligned} Z_\mathcal{M} &= \sum_{\mathbf{h} \in \mathbf{H}} \frac{\#(F_\mathbf{h} \wedge F_\mathcal{M}) \times \prod_{i=1}^{m} \phi_i(\mathbf{x}_{\mathbf{h},V(\phi_i)})}{U(\mathbf{h})} U(\mathbf{h}) \\ &= \mathbb{E}_U \left[ \frac{\#(F_\mathbf{h} \wedge F_\mathcal{M}) \times \prod_{i=1}^{m} \phi_i(\mathbf{x}_{\mathbf{h},V(\phi_i)})}{U(\mathbf{h})} \right] \end{aligned} \quad (12)$$

**Algorithm 2**: Formula Importance Sampling (FIS)

**Input**: A `PropMRF` $\mathcal{M}$ and an importance distribution $U(\mathbf{H})$ over a set of clauses $\mathbf{H} = \{H_1, \ldots, H_r\}$
**Output**: An unbiased estimate of $Z_\mathcal{M}$
**begin**
 $\widetilde{Z} = 0$, $N = 0$
 **repeat**
  $qb = 1$ (Backtrack-free probability is stored here),
  $G = F_\mathcal{M}$ and $\mathbf{h} = \phi$
  **for** $i = 1$ *to* $|\mathbf{H}|$ **do**
   Let $G_1 = G \wedge H_i$ and $G_0 = G \wedge \neg H_i$
   **if** $G_0$ *and* $G_1$ *have a solution (Checked using a SAT solver)* **then**
    Sample $h_i$ from $U(H_i|\mathbf{h})$
    $\mathbf{h} = \mathbf{h} \cup h_i$
    $qb = qb \times U(H_i = h_i|\mathbf{h})$
    $G = G \wedge (H_i = h_i)$
   **else**
    **if** $G_0$ *is Satisfiable* **then**
     $\mathbf{h} = \mathbf{h} \cup (H_i = 0)$
     $G = G \wedge (H_i = 0)$
    **else**
     $\mathbf{h} = \mathbf{h} \cup (H_i = 1)$
     $G = G \wedge (H_i = 1)$
  $w$ = sum of weights of soft clauses satisfied by $\mathbf{h}$
  $s$ = Estimate of model counts of $G$
  $\widetilde{Z} = \widetilde{Z} + s \times \exp(w)/qb$
  $N = N + 1$
 **until** *timeout*
 $\widetilde{Z} = \widetilde{Z}/N$
 **return** $\widetilde{Z}$
**end**

Given $N$ samples $\mathbf{h}^{(1)}, \ldots, \mathbf{h}^{(N)}$ generated from $U(\mathbf{H})$, we can estimate $Z_\mathcal{M}$ as $\widetilde{Z}_N$, where:

$$\widetilde{Z}_N = \frac{1}{N} \sum_{i=1}^{N} \frac{\#(F_{\mathbf{h}^{(i)}} \wedge F_\mathcal{M}) \times \prod_{j=1}^{m} \phi_j(\mathbf{x}_{\mathbf{h}^{(i)}, V(\phi_j)})}{U(\mathbf{h}^{(i)})} \quad (13)$$

There are two issues that need to be addressed in order to use Equation 13 for any practical purposes. First, the importance distribution $U(\mathbf{h})$ may suffer from the rejection problem (Gogate and Dechter, 2007a) in that we may generate truth assignments (to clauses) which are inconsistent, namely their model count is zero. Note that this could happen even if there are no hard clauses in $\mathcal{M}$ because the formula combinations considered may be inconsistent. Fortunately, if we ensure that $U(\mathbf{h}) = 0$ whenever $\mathbf{h}$ is inconsistent, namely make $U(\mathbf{h})$ backtrack-free (Gogate and Dechter, 2007a), we can avoid this problem altogether. Algorithm 2 outlines a procedure for constructing such a distribution using a complete SAT solver (for example Minisat (Sorensson and Een, 2005)). Second, computing $\#(F_{\mathbf{h}^{(j)}} \wedge F_\mathcal{M})$ exactly may be too time consuming. In such cases, we can use state-of-the-art approximate counting techniques such as ApproxCount (Wei and Selman, 2005), SampleCount (Gomes et al., 2007) and SampleSearch (Gogate and Dechter, 2007b).

### 4.3 Variance Reduction

The estimate $\widetilde{Z}_N$ output by Algorithm 2 is likely to have smaller mean squared error than $\widehat{Z}_N$ given in Equation 10. In particular, given a variable-based importance distribution $Q(\mathbf{X})$, we can always construct a formula based importance distribution $U(\mathbf{H})$ from $Q(\mathbf{X})$, such that the variance of $\widetilde{Z}_N$ is smaller than that of $\widehat{Z}_N$. Define:

$$U(\mathbf{h}) = \sum_{\mathbf{x_h} \in Sol(F_\mathbf{h} \wedge F_\mathcal{M})} Q(\mathbf{x_h}) \quad (14)$$

Intuitively, each sample from $U(\mathbf{H})$ given by Equation 14 is heavy in the sense that it corresponds to $\#(F_\mathcal{M} \wedge F_\mathbf{h})$ samples from $Q(\mathbf{x_h})$. Because of this larger sample size, the variance of $\widetilde{Z}_N$ is smaller than that of $\widehat{Z}_N$ (assuming that $\#(F_\mathcal{M} \wedge F_\mathbf{h})$ can be computed efficiently). The only caveat is that generating samples from $U(\mathbf{H})$ is more expensive. Formally (the proof is provided in the extended version of the paper available online),

THEOREM 1. *Given a* `PropMRF` $\mathcal{M}$*, a proposal distribution* $Q(\mathbf{X})$ *defined over the variables of* $\mathcal{M}$*, a set of formulas* $\mathbf{H} = \{H_1, \ldots, H_r\}$ *and a distribution* $U(\mathbf{H})$ *defined as in Equation 14, the variance of* $\widetilde{Z}_N$ *is less than or equal to that of* $\widehat{Z}_N$.

We can easily integrate FIS with other variance reduction schemes such as Rao-Blackwellisation (Casella and Robert, 1996) and AND/OR sampling (Gogate and Dechter, 2008). These combinations can lead to interesting time versus variance tradeoffs. We leave these improvements for future work. We describe how $U(\mathbf{H})$ can be constructed in practice in the next section.

## 5 Experiments

### 5.1 Exact Inference

We compared "Formula Decomposition and Conditioning (FDC)" against "Variable Decomposition and Conditioning (VDC)", variable elimination (VE) (Dechter, 1999) and ACE (Chavira and Darwiche, 2008) (which internally uses the C2D compiler (Darwiche, 2004)) for computing the partition function on benchmark problems from three domains: (a) Random networks, (b) medical diagnosis networks and (c) Relational networks. ACE, FDC and VDC use the same clausal representation while VE uses tabular representation. Note that the domains are deliberately chosen to elucidate the properties of FDC, in particular, to verify our intuition that as size of the clauses increases, FDC is likely to dominate VDC.

We implemented FDC and VDC on top of RELSAT (Roberto J. Bayardo Jr. and Pehoushek, 2000), which is a SAT model counting algorithm. As mentioned earlier, after conditioning on a formula, we use various Boolean propagation, pruning techniques such as unit propagation, clause learning, subsumption elimination and resolution. Also, similar to Cachet (Sang et al., 2005), we use component caching and similar to w-cutset conditioning (Dechter, 1999), we invoke bucket elimination at a node if the treewidth of the (remaining) PropMRF at the node is less than 16.

| Problem | w | FDC | VDC | ACE | VE |
|---------|------|---------|---------|--------|------|
| **Random** | | | | | |
| 40-40-3 | 10.80 | 0.08 | 0.08 | 0.60 | **0.01** |
| 40-40-5 | 23.00 | 11.81 | 10.57 | 108.37 | **1.69** |
| 40-40-7 | 29.80 | **11.77** | 245.63 | 13.37 | X |
| 40-40-9 | 33.40 | **1.71** | 326.42 | 14.37 | X |
| 50-50-3 | 12.60 | **0.02** | **0.02** | 0.73 | **0.02** |
| 50-50-5 | 28.40 | 278.25 | 257.45 | **56.94** | X |
| 50-50-7 | 36.00 | **167.79** | 1139.06 | 294.30 | X |
| 50-50-9 | 42.20 | **20.97** | 1187.28 | 113.52 | X |
| 60-60-3 | 15.00 | 0.08 | 0.11 | 0.58 | 0.04 |
| 60-60-5 | 33.80 | X | X | X | X |
| 60-60-7 | 44.00 | X | X | X | X |
| 60-60-9 | 49.60 | **218.28** | X | X | X |
| **QMRDT** | | | | | |
| 40-40-5 | 16.80 | 2.03 | 1.26 | 1.17 | **1.17** |
| 40-40-7 | 22.20 | 6.39 | 6.73 | 3.87 | **7.93** |
| 40-40-9 | 24.00 | 22.20 | 44.51 | **9.40** | X |
| 40-40-11 | 25.20 | 18.20 | 69.00 | **10.08** | X |
| 50-50-5 | 22.80 | 14.53 | 14.96 | 9.14 | **3.63** |
| 50-50-7 | 30.00 | 545.23 | 517.71 | **379.41** | X |
| 50-50-9 | 34.00 | **33.30** | 883.04 | 357.06 | X |
| 50-50-11 | 33.00 | **28.43** | 554.01 | 495.96 | X |
| 60-60-5 | 26.00 | 244.05 | **203.32** | 310.37 | X |
| 60-60-7 | 34.60 | **56.40** | 1096.62 | 637.23 | X |
| 60-60-9 | 40.40 | **97.20** | 1180.94 | 554.01 | X |
| 60-60-11 | 45.00 | **72.10** | X | 488.10 | X |
| **FS** | | | | | |
| fs-25-5 | 22.80 | 30.93 | 26.86 | 333.20 | **8.53** |
| fs-27-5 | 24.80 | 151.56 | **135.22** | 353.64 | X |
| fs-29-5 | 26.60 | 391.65 | 371.74 | **119.23** | X |
| fs-31-5 | 29.40 | 1312.90 | 892.20 | **357.65** | X |
| **Cora** | | | | | |
| Cora2 | 12.00 | 0.17 | 0.14 | 1.84 | **0.04** |
| Cora3 | 32.00 | **3902.20** | X | X | X |

Table 1: Average runtime in seconds of the four algorithms used in our study over 10 random instances for each problem. We gave each solver a time-bound of 3 hrs and a memory bound of 2GB. X indicates that either the memory or time bound was exceeded. The second column gives the average treewidth.

Since FDC is a DPLL-style backtracking search scheme, its performance is highly dependent upon a good branching heuristic (that selects the next clause to condition on). In our implementation, we used a simple dynamic heuristic of conditioning on the largest sub-clause (unit clause in case of VDC) that is common to most hard and soft clauses, ties broken arbitrarily. The main intuition for this heuristic is that branching on the largest common sub-clause would cause the most propagation, yielding the most reduction in the search space size. We also tried a few other heuristics, both static and dynamic, such as (i) conditioning on a sub-clause C (and its negation) that causes the most unit propagations (but one has to perform unit propagations for each candidate clause, which can be quite expensive in practice) (ii) graph partitioning heuristics based on the min-fill, min-degree and hmetis orderings; these heuristics are used by solvers such as ACE (Chavira and Darwiche, 2008) and AND/OR search (Dechter and Mateescu, 2007) and (iii) Entropy-based heuristics. The results for these heuristics show a similar trend as the results for the heuristic used in our experiments, with the latter performing better on an average. We leave the development of sophisticated formula-ordering heuristics for future work.

Table 1 shows the results. For each problem, we generated 10 random instances. For each instance, we set 5% of randomly chosen variables as evidence. Each row shows the average time in seconds for each problem.

### 5.1.1 Random networks

Our first domain is that of random networks. The networks are generated using the model $(n, m, s)$, where $n$ is the number of (Boolean) variables, $m$ is the number of weighted clauses and $s$ is the size of each weighted clause. Given $n$ variables $\mathbf{X} = \{X_1, \ldots, X_n\}$, each clause $C_i$ (for $i = 1$ to $m$) is generated by randomly selecting $s$ (distinct) random variables from $\mathbf{X}$ and negating each with probability $0.5$. For our experiments, we set $n = m$ and experimented with three values for $n$ and $m$: $n, m \in \{40, 50, 60\}$. $s$ was varied from 3 to 9 in increments of 2.

A random problem $(n, m, s)$ is designated as $n - m - s$ in Table 1. We see that FDC dominates VDC as $s$ increases. ACE is often inferior to FDC and often inferior to VDC. As expected, variable elimination which does not take advantage of the structure of the formulas is the fastest scheme when the treewidth is small but is unable to solve any problems having treewidth greater than 24.

### 5.1.2 Medical Diagnosis

Our second domain is a version of QMR-DT medical diagnosis networks (Shwe et al., 1991) as used in Cachet (Sang et al., 2005). Each problem can be specified using a two layer bipartite graph in which the top layer consists of diseases and the bottom layer consists of symptoms. If a disease causes a symptom, there is an edge from the disease to the symptom. We have a weighted unit clause for each disease and a weighted clause for each symptom, which is simply a logical OR of the diseases that cause it (in (Sang et al., 2005), this clause was hard. We attach an arbitrary weight to it to make the problem harder). For our experiments, we varied the numbers of diseases and symptoms from 40 to 60. For each symptom, we varied the number of diseases that can cause it from 5 to 11 in increments of 2. The diseases for each symptom are chosen randomly.

A QMR-DT problem $(d, f, s)$ is designated as $d - f - s$ in Table 1. We can see that as the size of the clauses increases,

FDC performs better than VDC. FDC also dominates ACE as the problem size increases.

### 5.1.3 Relational networks

Our final domain is that of relational networks. We experimented with the Friends and Smokers networks and the Entity resolution networks.

In the friends and smokers networks (FS), we have three first order predicates $smokes(x)$, which indicates whether a person smokes, $cancer(x)$, which indicates whether a person has cancer, and $friends(x, y)$, which indicates who are friends of whom. The probabilistic model is defined by assigning weights to two logical constraints, $friends(x, y) \land smokes(x) \Rightarrow smokes(y)$ and $smokes(x) \Rightarrow cancer(x)$. Given a domain for $x$ and $y$, a PropMRF can be generated from these two logical constraints by considering all possible groundings of each predicate. We experimented with different domain sizes for $x$ and $y$ ranging from 25 to 34. From Table 1, we can see that the time required by FDC is almost the same as VDC. This is because the size of the clauses is small ($\leq 3$). ACE dominates both FDC and VDC.

Entity resolution is the problem of determining which observations correspond to the same entity. In our experiments, we consider the problem of matching citations of scientific papers. We used the CORA Markov logic network given in the Alchemy tutorial (Kok et al., 2004). This MLN has ten predicates such as $Author(bib, author)$, $Title(bib, title)$, $SameAuthor(author, author)$, $SameTitle(title, title)$ etc. and clauses ranging from size 2 to 6. The clauses express relationship such as: if two fields have high similarity, then they are (probably) the same; if two records are the same, their fields are the same, and vice-versa; etc. We experimented with domain sizes of 2 and 3 for each of the 5 first-order variables present in the domain. The problems are denoted as cora2 and cora3 respectively. From Table 1, we can see that FDC is the only algorithm capable of solving the largest instance.

### 5.2 Approximate Inference

We compared "Formula importance sampling (FIS)" against "Variable importance sampling (VIS)" and state-of-the-art schemes such as MC-SAT (Poon and Domingos, 2006) and Gibbs sampling available in Alchemy (Kok et al., 2004) on the three domains described above. For both VIS and FIS, we chose to construct the importance distribution $Q$ from the output of a Belief propagation scheme (BP), because BP was shown to yield a better importance function than other approaches in previous studies (Yuan and Druzdzel, 2006; Gogate and Dechter, 2005).

We describe next, how the method described in (Gogate and Dechter, 2005) can be adapted to construct an importance distribution over formulas. Here, we first run BP (or Generalized Belief Propagation (Yedidia et al., 2004)) over a factor (or region) graph in which the nodes are the variables and the factors are the hard and the soft clauses. Let $(C_1, \ldots, C_m)$ be an ordering over the soft clauses. Given a truth assignment to the first $i - 1$ soft clauses $\mathbf{c}_{i-1} = (c_1, \ldots, c_{i-1})$, we compute $U(C_i|\mathbf{c}_{i-1})$ as follows. We first simplify the formula $F = F_\mathcal{M} \land F_{\mathbf{c}_{i-1}}$, possibly deriving new unit clauses. Let $\phi_{C_i}$ be the marginal distribution at the factor corresponding to the clause $C_i$ in the output of BP. Then, $U(C_i|\mathbf{c}_{i-1})$ is given by:

$$U(C_i = True|\mathbf{c}_{i-1}) \propto \sum_{\mathbf{y} \in \phi_{C_i}} I_{F,C_i}(\mathbf{y}) \phi_{C_i}(\mathbf{y}) \quad (15)$$

where $I_{F,C_i}(\mathbf{y}) = 1$ if $\mathbf{y}$ evaluates $C_i$ to True but does not violate any unit clause in $F$, and 0 otherwise. Note that the importance distribution $Q$ over the variables is a special case of the scheme described above in which we construct a distribution over all the unit clauses.

We implemented Algorithm 2 as follows. Notice that the algorithm requires a SAT solver and a model counter. We used Minisat (Sorensson and Een, 2005) as our SAT solver. For model counting, we use the RELSAT model counter whenever exact counting was feasible[2] and the approximate solver SampleSearch (Gogate and Dechter, 2007b) whenever it wasn't.

We measure the performance of the sampling schemes using the sum Kullback-Leibler divergence (KLD) between the exact and the approximate posterior marginals for each variable given evidence. Time versus sum KLD plots for two representative problems from each domain are shown in Figures 4, 5 and 6. We can clearly see that as the size of the clauses increases, FIS outperforms VIS, MC-SAT and Gibbs sampling.

## 6 Summary and Conclusion

In this paper, we introduced a new formula-based approach for performing exact and approximate inference in graphical models. Formula-based inference is attractive because: (a) it generalizes standard variable-based inference, (b) it yields several new efficient algorithms that are not possible by reasoning just over the variables and (c) it fits naturally within the recent research efforts in combining logical and probabilistic Artificial Intelligence.

Our empirical evaluation shows that formula-based approach is especially suitable for domains having large clauses. Such clauses are one of the main reasons for using logic instead of tables for representing potentials or

---

[2] Exact counting was invoked if the number of variables was less than 100, which was the case for most networks that we experimented with, except the relational benchmarks.

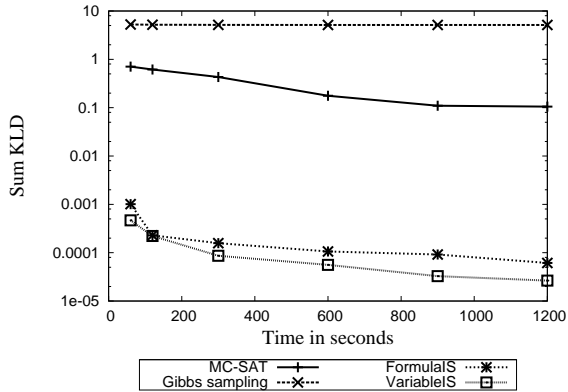
(a) Random problem ($n = 50, m = 50, s = 5$)

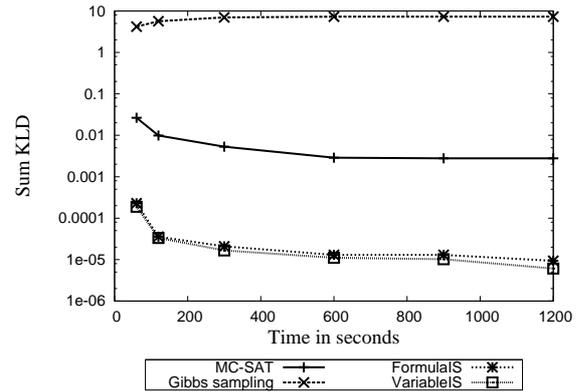
(a) QMR-DT problem ($d = 50, f = 50, s = 5$)

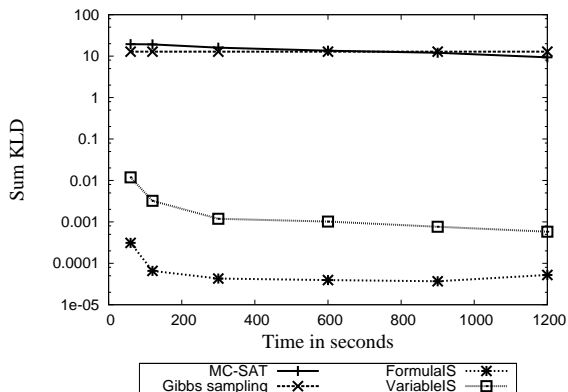
(b) Random problem ($n = 50, m = 50, s = 7$)

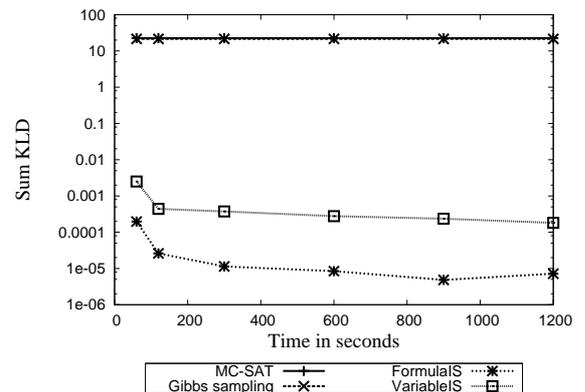
(b) QMR-DT problem ($d = 50, f = 50, s = 11$)

Figure 4: Time versus Sum KLD plots for 2 Random instances.

Figure 5: Time versus Sum KLD plots for 2 QMR-DT networks.

CPTs in graphical models. In particular, conventional tabular representations require space exponential in the number of variables in the scope of the potential, while if the potential can be summarized using a constant number of clauses, we only require linear space. Since an efficient inference scheme is one of the main bottleneck in learning PropMRFs having large clauses, we believe that our formula-based approach to inference can lead to new structure and weight learning schemes that learn large weighted clauses from data.

Our work can be extended in several ways. In particular, we envision formula-based versions of various inference schemes such as variable elimination, belief propagation and Markov Chain Monte Carlo (MCMC) sampling. One of these schemes, namely formula elimination trivially follows from this work, as it is known that conditioning works along the reverse direction of elimination (Dechter, 1999). Also, we envision the development of lifted versions of all the formula-based schemes proposed in this paper.

**Acknowledgements**

This research was partly funded by ARO grant W911NF-08-1-0242, AFRL contract FA8750-09-C-0181, DARPA contracts FA8750-05-2-0283, FA8750-07-D-0185, HR0011-06-C-0025, HR0011-07-C-0060 and NBCH-D030010, NSF grants IIS-0534881 and IIS-0803481, and ONR grant N00014-08-1-0670. The views and conclusions contained in this document are those of the authors and should not be interpreted as necessarily representing the official policies, either expressed or implied, of ARO, DARPA, NSF, ONR, or the United States Government.

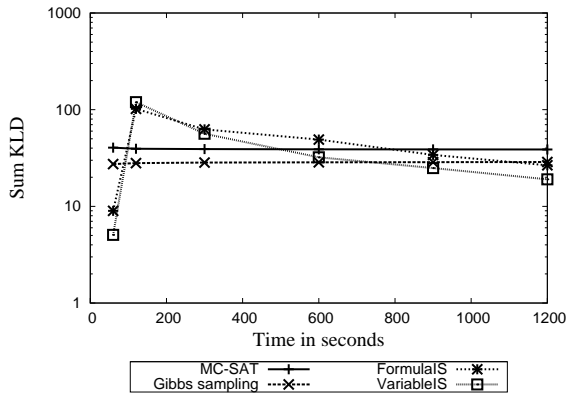

(a) Friends and Smokers network over 27 individuals

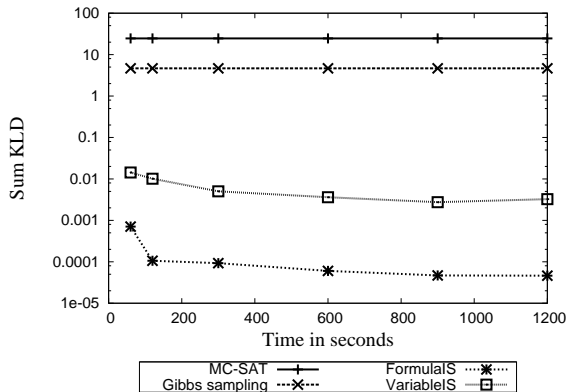

(b) CORA Markov Logic network with 3 Authors, Venues, Titles, Words and Bibliographies

Figure 6: Time versus KLD plots for Relational instances.